\documentclass{article}
\usepackage[utf8]{inputenc}
\usepackage{enumitem}
\usepackage{graphicx}
\usepackage{tikz}
\usepackage{xspace}
\usepackage{subcaption}
\usepackage{xcolor}
\usepackage{float}
\usepackage{placeins}
\usepackage{indentfirst,csquotes}
\usepackage{newunicodechar}
\newunicodechar{̆}{\u{}}
\usepackage[backend=biber,style=numeric, sorting=none]{biblatex}
\usepackage{amsmath}
\usepackage{amssymb}
\usepackage{mathtools}
\usepackage{amsthm}
\usepackage[table]{xcolor}
\usepackage{microtype}
\usepackage{graphicx}
\usepackage{subcaption}
\usepackage{booktabs} % for professional tables
\usepackage{enumitem}
\usepackage{amssymb,amsthm,amsmath}
\usepackage{xcolor,hyperref,titlesec,fancyhdr,etoolbox}

\newcommand{\ourmethod}{\textsc{HetScene}\xspace}

\providecommand{\Description}[1]{}
\definecolor{titlepurple}{HTML}{B66CD1}
\definecolor{colorfirst}{RGB}{255, 204, 204}
\definecolor{colorsecond}{RGB}{255, 230, 204}
\definecolor{colorthird}{RGB}{255, 251, 214}
\newcommand{\best}{{\cellcolor{colorfirst}}}
\newcommand{\second}{{\cellcolor{colorsecond}}}
\newcolumntype{N}{>{\centering\arraybackslash}p{3em}}
\titleformat{\section}[block]
  {\normalfont\large\bfseries} % \large 比 \Large 小一级，比正文稍微大一点且加粗
  {\thesection.}               % 编号后面加个点
  {0.6em}                      % 编号与标题文字的间距
  {}

\titlespacing*{\section}{0pt}{2ex plus 1ex minus .2ex}{1.5ex plus .2ex}

\hypersetup{ colorlinks=true, linkcolor=black, filecolor=black, urlcolor=black }

\usepackage{lipsum}

\begin{document}
\title{\textcolor{titlepurple}{\textsc{HetScene}}: \textcolor{titlepurple}{Het}erogeneity-Aware Diffusion for Dense Indoor \textcolor{titlepurple}{Scene} Generation}
\author{Zini Chen, Junming Huang, Rong Zhang, \\Jiamin Xu, Cheng Peng, Chi Wang, Weiwei Xu}
\date{\today}
% \address{Address}
% \email{example@mail.com}
\maketitle

\begin{figure}[H]
  \centering
  \includegraphics[width=\linewidth]{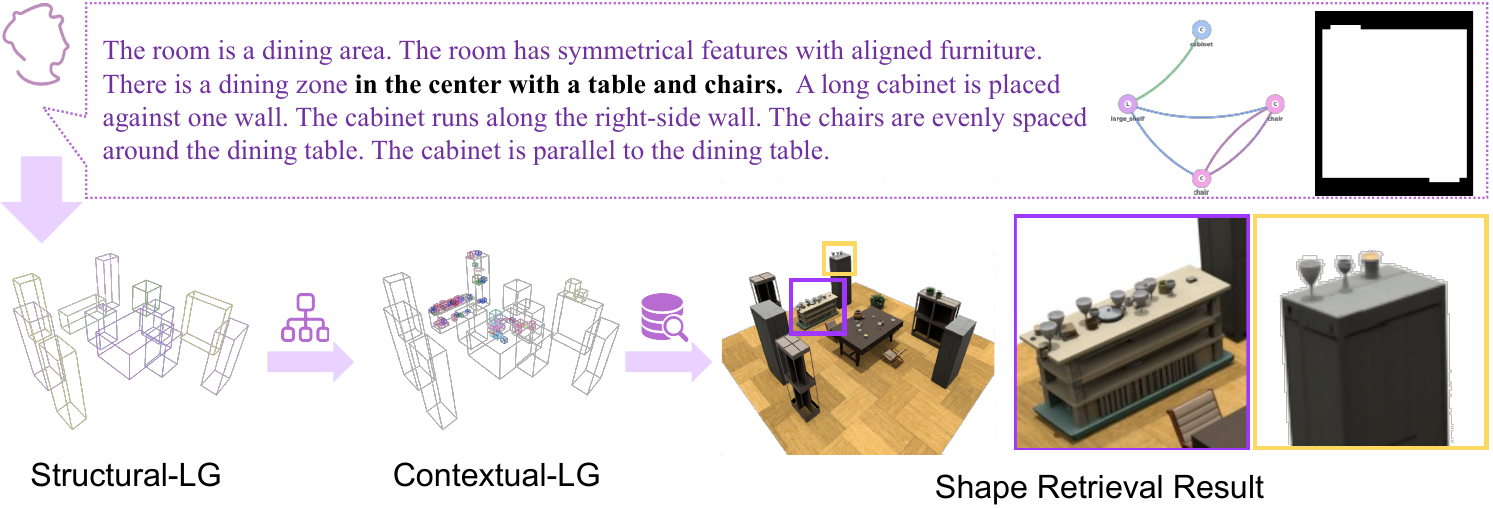}
  % \vspace{-0.7cm}
  \caption{Given a text prompt, a relation graph, and a room footprint as input,
  \ourmethod generates a dense indoor layout in two stages.
  \textit{Structural Layout Generation} captures the global functional
  organization of the room by synthesizing primary objects, while
  \textit{Contextual Layout Generation} takes the generated primary
  layout as spatial context and support guidance, and places secondary
  objects around their corresponding primary objects.
  The resulting layout is finally instantiated into a complete indoor
  scene by retrieving 3D assets from an asset collection.}
  \label{fig:teaser}
\end{figure}

\let\thefootnote\relax

\begin{abstract}
    Generating controllable and physically plausible indoor scenes is a pivotal prerequisite for constructing high-fidelity simulation environments for embodied AI. However, existing deep learning-based methods usually treat all objects as homogeneous instances within a unified generation process. While effective for sparse and simplistic layouts, they struggle to model realistic layouts with dense object arrangements and complex spatial dependencies, leading to limited scalability and degraded physical plausibility.
    To deal with these challenges, we revisit indoor layout generation from the perspective of structural heterogeneity and decompose the objects into primary objects and secondary objects according to their distinct roles in shaping a scene. Based on this decomposition, we propose \ourmethod, a heterogeneous two-stage generation framework that decouples indoor layout synthesis into Structural Layout Generation (SLG) and Contextual Layout Generation (CLG).
    SLG first generates globally coherent structural layouts with only primary objects conditioned on text descriptions, top-down binary room masks, and spatial relation graphs, establishing a stable global macro-skeleton of large core furniture. Then CLG takes the generated structural layout as support geometric anchors to model the cross-scale long-range dependencies and synthesize dense secondary objects through a local conditional generation network with a carefully-designed learnable spatial-semantic modulation mechanism. By aligning each generation stage with the spatial and semantic regularities of different object groups, our method reduces the entanglement between global room planning and local support-aware placement. Experimental results on the challenging multi-source M3DLayout benchmark demonstrate that \ourmethod achieves state-of-the-art performance, enabling scalable generation of realistic, physically plausible, and high-density indoor scenes.
\end{abstract}

\section{Introduction}
% Realistic 3D indoor layout generation has become an important research topic shared by computer graphics and embodied AI. As artificial intelligence gradually moves from pattern recognition in digital spaces toward embodied interaction in the physical world, reliable indoor scene generation has become a key piece of infrastructure for that shift.
Realistic 3D indoor layout generation is a research topic of growing importance at the intersection of computer graphics and embodied AI, serving as critical infrastructure for training and evaluating embodied agents in simulated physical environments.
% However, when layouts involve dense object arrangements or complex spatial relationships that must be satisfied jointly, existing approaches remain limited in both layout quality and scalability.
%Generating realistic 3D indoor layouts is a fundamental problem for scene synthesis, virtual environment creation, and embodied AI. 

Early approaches typically formulate this task as a procedural or optimization problem driven by rule-based constraints, handcrafted heuristics, or scene priors~\cite{yehSynthesizingOpenWorlds2012,qiHumancentricIndoorScene2018,raistrickInfinigenIndoorsPhotorealistic2024,yuMakeItHome2011,merrellInteractiveFurnitureLayout2011}. Although such methods provide physical plausibility, their expressive capacity is inherently limited by manually designed heuristics. They often struggle to capture the complex semantic relationships, diverse spatial configurations, and long-tail object distributions in real-world scenes, leading to stylistically monotonous generation results.

%Procedural and optimization-based methods for indoor layout generation typically formulate this task as an optimization problem driven by rule-based constraints, handcrafted heuristics, or scene priors\cite{yehSynthesizingOpenWorlds2012,qiHumancentricIndoorScene2018,raistrickInfinigenIndoorsPhotorealistic2024,yuMakeItHome2011,merrellInteractiveFurnitureLayout2011}. Such methods have inherent advantages in ensuring physical plausibility. However, their expressive capacity heavily depends on manually designed heuristics, making it difficult to exhaustively model the complex semantic relationships and long-tail distributions among objects in real-world scenes. This leads to two major limitations. First, the generated layouts are often stylistically monotonous and lack diversity and creativity. Second, the optimization process is computationally expensive and struggles to provide fine-grained control over topological relationships among objects, thereby limiting the controllability and generalization ability of these methods in complex scenes. %后半部分花太多笔墨在非sota旧方法上了

To overcome these limits, recent works have shifted toward data-driven 3D layout generation, where auto-regressive models or diffusion models are adopted to learn the object placement distributions directly from large-scale scene datasets~\cite{paschalidouATISSAutoregressiveTransformers2021, tangDiffuSceneDenoisingDiffusion2024a, huangBuildingBlockHybridApproach2025, huMixedDiffusion3D2024}. These methods have significantly improved the diversity and realism of generated layouts. Nevertheless, most progress is achieved under simplified settings, where scenes are relatively sparse, small objects are filtered and object counts are carefully controlled. This strategy improves the tractability but also removes the challenging structures. When it comes to realistic indoor environments, layouts involve dense object arrangements and complex spatial relationships that must be satisfied jointly. In these circumstances, existing data-driven approaches often demonstrate limited capabilities in producing high-quality and scalable layouts. 
% have significantly improved the diversity and realism of layout generation by learning object placement patterns and semantic relationships from large-scale data
% However, current progress has mainly been achieved in simple and sparse indoor scenes. % More details？ such as "with no more than xxx obj"
%To work around this, mainstream pipelines preprocess the training data by filtering out small objects and capping the per-room object count, which sidesteps rather than resolves the underlying difficulty.

Beyond the data limitation, we further observe that the degradation may stem from the modeling mismatch between existing generative formulations and the heterogeneous composition of indoor scenes.
%This degradation may stem from two factors. First, existing 3D scene datasets have limited data scale and provide insufficient coverage of complex layouts. Second, 
Current generative pipelines typically adopt a homogeneous formulation, treating all objects as instances to be generated within the same modeling process, which neglects the heterogeneity that different objects play distinct roles in providing spatial and semantic information that shapes an indoor scene. Specifically, in a typical room, objects can be broadly divided into two groups that contribute to the layout in different ways. 
The first is \textit{primary objects}, such as beds, sofas, tables, and cabinets. These objects usually occupy substantial floor area, are relatively independent of one another, and are mainly governed by global room-level constraints, including wall alignment, functional zones, accessibility, and walkable space. The second group is \textit{secondary objects}, such as books, desk lamps, pillows, monitors, and decorations. These objects are typically smaller, more numerous, and strongly dependent on nearby primary objects through physical support, contact, and semantic co-occurrence. These two groups follow distinct statistical and geometric distributions: primary objects are sparse and globally constrained, whereas secondary objects are dense and locally conditioned.

Mainstream diffusion-based layout models conflate these two regimes. DiffuScene~\cite{tangDiffuSceneDenoisingDiffusion2024a} and MiDiffusion~\cite{huMixedDiffusion3D2024} represent all objects in a room as a unified, fixed-length object attributes and jointly denoise them within a shared representation space, forcing the network to simultaneously handle global room planning for primary objects and local support-aware placement for secondary objects.
BuildingBlock~\cite{huangBuildingBlockHybridApproach2025} introduces a hierarchical modeling process at the pipeline level, but it still generates all objects in a unified representation, while fine-grained objects are adjusted through post-processing.
Without explicit modeling of support relations and functional co-occurrence priors, the dense local dependencies introduced by secondary objects rapidly increase the generation complexity, making existing layout generation models struggle to scale to realistic, object-rich indoor scenes.
%BuildingBlock~\cite{huangBuildingBlockHybridApproach2025} introduces hierarchy at the pipeline level, but its diffusion process still operates on a flat object representation, with structure injected only through post-processing. %% Flat obj representation ??
%Without explicit modeling of support relations or co-occurrence priors, the combinatorial growth introduced by secondary objects rapidly overwhelms such unified denoisers, which explains why these models remain robust only on filtered, sparse datasets. %%% somehow strange, need polish 

%This diagnosis is consistent with complementary evidence from manipulation-oriented work~\cite{liuStructFormerLearningSpatial2021, liuStructDiffusionLanguageGuidedCreation2023, jiaClutterGenClutteredScene2024, muraliCabiNetScalingNeural2023}, in which data-driven models successfully arrange dense small objects on bounded supports by exploiting local contact cues. Their success indicates that learned models are well-suited to secondary objects when the task is properly localized, and that the bottleneck of current scene-level methods lies not in data-driven modeling itself, but in conflating two regimes that demand different inductive biases.

Motivated by this observation, we reformulate indoor layout generation as a heterogeneous generation problem that explicitly aligns the generative process with the structural roles of different object groups and propose \ourmethod, a two-stage diffusion-based framework for dense indoor scene generation.
In the first stage, \textit{Structural Layout Generation} (SLG) establishes the global functional organization of the room by generating primary objects conditioned on text descriptions, room masks, and spatial relation graphs. In the second stage, \textit{Contextual Layout Generation} (CLG) takes the generated primary layout as spatial context and support guidance, and synthesizes secondary objects around their corresponding primary objects. 
This decomposition transforms the original problem from joint generation over all objects into a combination of global layout planning and local support-conditioned object placement.
By assigning each object group to a generation process that matches its spatial and semantic regularities, our framework allows each stage to model a better-conditioned sub-distribution: sparse, globally constrained primary layouts in SLG, and dense, locally conditioned secondary arrangements in CLG. In this way, our framework can improve the plausibility and scalability of object-rich indoor layout generation without imposing a fixed upper bound on the number of objects per room. 
Our contributions are summarized as follows:
\begin{itemize}[leftmargin=*, itemsep=2pt, topsep=2pt]
    \item
    We propose \ourmethod, a heterogeneity-aware layout generation framework that decomposes indoor scene synthesis into structural layout generation and context-conditioned layout generation, enabling global room organization and local support-aware placement to be modeled separately. The CLG stage can serve as a plug-and-play component compatible with arbitrary upstream coarse layouts.
    %and can be utilized to refine the generated layouts iteratively.
    
    %We decompose indoor layout generation into SLG and CLG, reducing the entanglement between global structure planning and local object placement. The refinement module uses generated large objects as support priors to synthesize small objects, and can serve as a plug-and-play component compatible with arbitrary upstream coarse layouts.

    \item
    We introduce a learnable spatial-semantic modulation mechanism that adaptively calibrates the relative contribution of continuous geometric attributes and discrete semantic embeddings during layout token construction, enabling stable joint training of primary and secondary objects in dense scenes. This calibration is empirically necessary for convergence under high object density in the second stage.
    %We introduce a unified conditional generation scheme that integrates text descriptions, room masks, spatial relation graphs, and generated large-object poses to improve controllability, support awareness, and physical plausibility. 
    % We unify text descriptions, room masks, and spatial relation graphs within a single conditional generation framework. In the second stage, large-object poses are further introduced as structural conditions, turning small-object synthesis from global sampling into local conditional generation and improving controllability and physical plausibility.

    \item
    Extensive experiments show that our method enables stable generation of realistic, high-density indoor scenes with dozens to hundreds of objects, alleviating small-object omission and capacity limitations of single-stage methods.
    
    %By introducing inductive biases for support relationships and object co-occurrence, our framework mitigates the capacity limitations and small-object omission issues of single-stage methods. It enables stable generation of dense indoor scenes containing dozens to hundreds of objects, advancing data-driven layout generation toward realistic, high-density environments.
\end{itemize}

\section{Related Work}
\subsection{Procedural Scene Generation}

\begin{figure*}[htp] 
    \centering
    \includegraphics[width=0.95\textwidth]{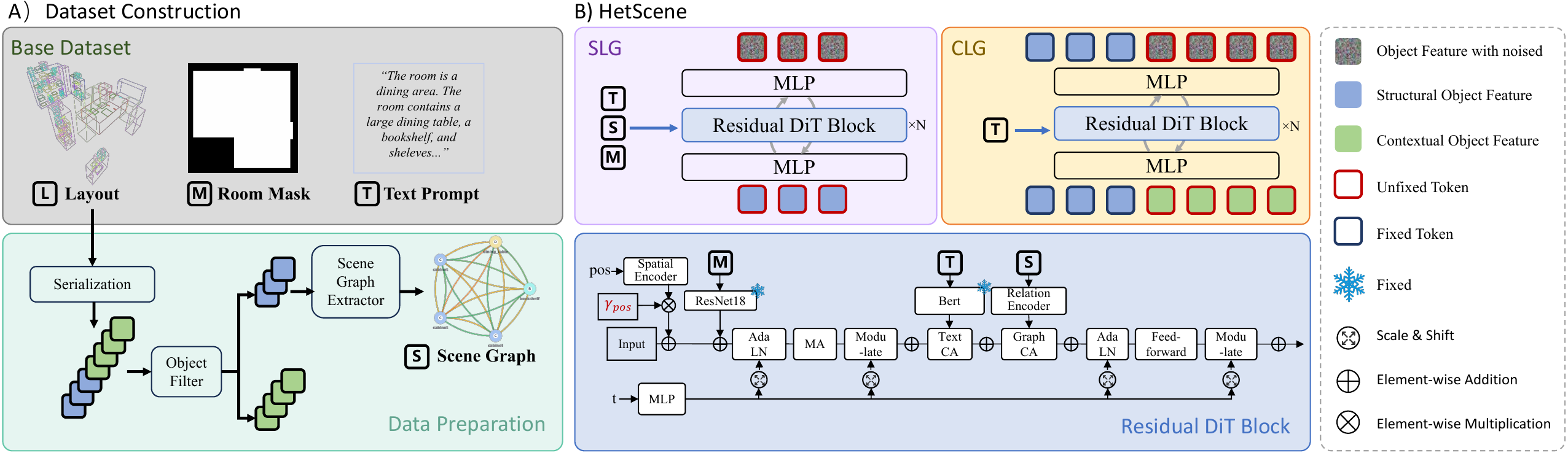}
    \caption{The core pipeline of HetScene.
A) Dataset Construction unveils our data representation and
the corresponding processing pipeline.
B) HetScene unveils the network architecture as well as the
input/output formulation of the model.}
    \label{fig:pipeline}
    \Description{}
  \vspace{-2mm}
    
\end{figure*}

Early indoor scene synthesis methods mainly relied on procedural rules or optimization-based pipelines~\cite{yehSynthesizingOpenWorlds2012,qiHumancentricIndoorScene2018,yuMakeItHome2011,merrellInteractiveFurnitureLayout2011}. Rule- and grammar-based systems construct rooms by composing predefined object templates according to manually designed production rules. Other methods formulate indoor layout design as an optimization problem, where object placements are guided by style preferences and spatial relations, and are typically solved with strategies such as simulated annealing or Markov Chain Monte Carlo (MCMC). More recently, systems such as Infinigen-Indoors~\cite{raistrickInfinigenIndoorsPhotorealistic2024} have extended this line of work to large-scale, fully procedural indoor environments with rich geometry and appearance variation.
These methods are effective at producing clean and physically plausible layouts, and they remain useful when large-scale training data is unavailable. However, their quality and diversity largely depend on hand-crafted rules, object libraries, and carefully tuned constraints. As a result, they are difficult to scale to open-ended object categories and struggle to capture the diverse styles and long-tail arrangements found in real indoor scenes. This limitation has motivated a shift toward data-driven methods that learn scene distributions directly from examples.

\subsection{Learning-Based Indoor Scene Synthesis}
Learning-based methods aim to generate indoor layouts by modeling scene distributions from data. Early works explored autoregressive models and graph neural networks to synthesize compact furniture arrangements. 
These methods~\cite{paschalidouATISSAutoregressiveTransformers2021, wangSceneFormerIndoorScene2021, gaoSceneHGNHierarchicalGraph2023, dhamoGraphto3DEndtoEndGeneration2021, wangPlanITPlanningInstantiating2019} learn object categories and spatial configurations from existing 3D scene datasets~\cite{zhangM3DLayoutMultiSourceDataset2026, fu3DFRONT3DFurnished2021, changMatterport3DLearningRGBD2017}, enabling more diverse and realistic layouts than purely rule-based approaches.
More recently, diffusion-based methods have become a dominant direction for indoor scene synthesis. These methods~\cite{tangDiffuSceneDenoisingDiffusion2024a, linInstructSceneInstructionDriven3D2024, yangPhyScenePhysicallyInteractable2024, huMixedDiffusion3D2024, zhaiEchoSceneIndoorScene2025, weiLEGONetLearningRegular2023,huangBuildingBlockHybridApproach2025} improve generation quality and controllability by conditioning on text descriptions, room geometry, scene graphs, or physical constraints. Despite these advances, most existing learning-based methods still represent a scene as a flat sequence or set of object tokens, and generate all objects within a single shared representation space. This formulation entangles room-level functional planning with object-level placement, making it difficult to model dense scenes with many cross-scale objects. In practice, many benchmarks and preprocessing pipelines restrict the maximum number of objects or remove small items, which limits their ability to generate complete indoor scenes. When these restrictions are relaxed, the lack of explicit support modeling often leads to missing small objects, weak local coherence, floating artifacts, and object penetration.

\subsection{Small-Object Arrangement}
Another line of work focuses on arranging small objects on bounded support surfaces, often motivated by robotic manipulation, rearrangement, and grasping. StructFormer~\cite{liuStructFormerLearningSpatial2021} and StructDiffusion~\cite{liuStructDiffusionLanguageGuidedCreation2023} use language instructions to organize tabletop objects into structured configurations, while ClutterGen~\cite{jiaClutterGenClutteredScene2024} and CabiNet~\cite{muraliCabiNetScalingNeural2023} generate dense cluttered scenes for grasping, collision reasoning, or manipulation planning. These methods provide useful inductive biases for modeling support relations, contact, and physical feasibility.
However, their scope is usually limited to a single tabletop, shelf, container, or local workspace. They typically do not take a full room layout as context, nor do they model how small objects should be distributed across multiple supporting furniture pieces in a complete indoor scene. This leaves a gap between room-level layout synthesis, which often ignores small objects, and tabletop arrangement, which abstracts away the surrounding room.

\section{Method}
This section presents \ourmethod, a two-stage hierarchical diffusion framework for text-driven structured indoor scene synthesis. \ourmethod uses a heterogeneity-aware design with two stages: \emph{Structural Layout Generation} (SLG) for primary objects and \emph{Structural Layout Generation} (CLG) for secondary objects conditioned on the SLG. The framework is shown in Fig.\ref{fig:pipeline}.

\subsection{Preliminary}

\paragraph{Denoising Diffusion Probabilistic Models (DDPMs)}
DDPMs~\cite{ho2020denoising} model the data distribution by learning to reverse a forward Markov process~\cite{dynkin1965markov} that progressively corrupts data with Gaussian noise.
Given a clean sample $\mathbf{x}_0$, the forward process is defined as
\begin{equation}
  q(\mathbf{x}_t \mid \mathbf{x}_0) = \mathcal{N}\!\left(\mathbf{x}_t;\,\sqrt{\bar\alpha_t}\,\mathbf{x}_0,\,(1-\bar\alpha_t)\mathbf{I}\right),
\end{equation}
where $\bar\alpha_t = \prod_{s=1}^{t}(1-\beta_s)$ and $\beta_t $ is a predefined noise variance schedule.
The reverse denoising process is learned by a network $\epsilon_\theta(\mathbf{x}_t, t)$ via the objective:
\begin{equation}
  \mathcal{L}_{\text{reverse}} = \mathbb{E}_{\mathbf{x}_0,\,\epsilon,\,t}\!\left[\,\bigl\|\epsilon - \epsilon_\theta(\mathbf{x}_t, t)\bigr\|^2\right].
\end{equation}

\paragraph{DiffuScene}
DiffuScene targets 3D indoor scene layout generation with a denoising network whose architecture follows an MLP encoder, a 1D-Unet backbone, and an MLP decoder.
The backbone operates on a slot sequence of per-object attributes and captures inter-object spatial dependencies through multi-level self-attention.
The network adopts $v$-prediction parameterization, predicting $v_t = \sqrt{\bar\alpha_t}\,\epsilon - \sqrt{1-\bar\alpha_t}\,\mathbf{x}_0$, which stabilizes training at low noise levels.
Training jointly optimizes a noise prediction loss and an IoU auxiliary loss:
\begin{equation}
  \mathcal{L} = \mathbb{E}_{\mathbf{X}_0,\,\epsilon,\,t}\!\left[\bigl\|\epsilon - \epsilon_\theta(\mathbf{X}_t, t)\bigr\|^2\right]
              + \lambda\,\mathcal{L}_{\text{IoU}},
\end{equation}
where $\mathcal{L}_{\text{IoU}}$ penalizes pairwise bounding box overlaps to suppress object collisions in the predicted layout.

\subsection{Transformer-based Denoising Backbone}
To represent a scene as an unordered set of object attributes, we address the limitation of the 1D-UNet backbone used in DiffuScene, which introduces task-irrelevant sequential order bias into the representation. At the structural level, we draw inspiration from BuildingBlock~\cite{huangBuildingBlockHybridApproach2025} and replace the denoising network with a Point·E-style Transformer~\cite{nichol2022point}, utilizing AdaLN from DiT~\cite{peebles2023scalable} to inject diffusion timestep conditions. Furthermore, rather than employing traditional positional encodings, spatial geometric cues are provided via spatial encodings from the spatial positions of object bounding boxes. Under this configuration, each object token is composed of multi-channel attribute linear embeddings integrated with the aforementioned spatial terms.

% \paragraph{Balanced Spatial-Semantic Embedding.} 
\paragraph{Learnable Spatial-semantic Modulation Mechanism}
The naive linear summation of spatial and semantic embeddings adopted by recent spatial-encoding layout backbones~\cite{huangBuildingBlockHybridApproach2025} causes the numerically larger geometric signals to overwhelm the categorical branch and prevents convergence in dense scenes. To address this, we depart from the summation scheme used in BuildingBlock and introduce an adaptive modulation mechanism. Specifically, we employ a learnable scalar $\gamma_{pos}$ to weight the injection of spatial embeddings into the primary features $x$, formally expressed as:
% To balance the relative contributions of geometric information and semantic features within the representation space, we depart from the simple linear summation scheme used in BuildingBlock~\cite{huangBuildingBlockHybridApproach2025} and introduce an adaptive modulation mechanism. Specifically, we employ a learnable scalar $\gamma_{pos}$ to weight the injection of spatial embeddings into the primary features $x$ , which is formally expressed as follows:
\begin{equation}
x' = x + \gamma_{pos} \cdot e_{pos}
\end{equation}
where $e_{pos}$ represent the spatial embeddings. By initializing $\gamma_{pos}$ to a minimal value, we effectively prevent the spatial signals from numerically overwhelming discrete semantic attributes during the early stages of training, thereby ensuring balanced optimization across multimodal features. Empirically, we find this single-scalar calibration to be a prerequisite for the convergence of the secondary-object branch under high object density.

\subsection{Conditional Representations}

\paragraph{Learnable Scene Graph (LSG) condition.}

While self-attention mechanisms are capable of implicitly capturing inter-object dependencies from data, they remain limited in modeling explicit topological constraints and spatial relationships. To address this, we propose the Learnable Scene Graph (LSG) condition, which explicitly encodes a relationship graph $G = (V, E)$ as a conditional signal. 

In $G$, $V$ represents the set of vertices where each vertex $v_i$ is defined by a discrete semantic attribute $c_i$ and an intra-class index $k_i \in \{0, 1, \dots\}$ to distinguish between instances of the same category. $E$ denotes the set of directed edges that define spatial constraints between objects, with specific relationship categories detailed in the supplementary material. The relationship tensor is strictly aligned with the object indices in the layout to ensure that the encoder maps each endpoint to its corresponding semantic attributes rather than an arbitrary sequence position.

To represent directed relationships while precisely distinguishing among multiple instances within the same category, we implement independent parameterization schemes for source and destination nodes. The essence of this architectural design is the direct mapping of discrete semantic labels and instance indices into continuous embedding spaces. Through end-to-end optimization, the model transcends the limitations of predefined geometric rules to autonomously capture latent spatial cues that facilitate coherent scene layout generation. Given the attributes $(c_i, k_i)$ and $(c_j, k_j)$ for instances $i$ and $j$ respectively, the directed endpoint embeddings are derived by aggregating learnable category and instance embeddings. Subsequently, the source embedding $e_i^{\text{src}}$, destination embedding $e_j^{\text{dst}}$, and relationship embedding $e_j^{\text{dst}}$ are concatenated and processed through a Multi-Layer Perceptron (MLP) to yield the initial edge features $h_{ij} \in \mathbb{R}^{d_h}$, which can be formulated as follows:
\begin{equation}
e_i^{\text{src}} = E_c^{\text{src}}(c_i) + E_k^{\text{src}}(k_i), \quad e_j^{\text{dst}} = E_c^{\text{dst}}(c_j) + E_k^{\text{dst}}(k_j),
\end{equation}
\begin{equation}
h_{ij} = \text{MLP}([e_i^{\text{src}} \parallel e_j^{\text{dst}} \parallel  e_j^{\text{dst}}]),
\end{equation}
In these formulations, the $\text{MLP}$ utilizes the GELU activation function, $\parallel$ denotes the vector concatenation operation, and $d_h$ represents the hidden dimension of the backbone network.

Upon obtaining the representations for all valid edges $\{h_{ij}\}$, they are fed into a lightweight Transformer encoder. Self-attention is applied across these edge features to allow the representation of each edge to fuse with the context of others, thereby modeling the interactions and synergies among multiple relational constraints. Finally, to achieve a balance between computational overhead and constraint intensity, the encoded edge features undergo masked mean pooling to be aggregated into a single global graph token $g \in \mathbb{R}^{d_h}$. This token serves as the context for cross-attention to realize conditional guidance for the global spatial layout.

\paragraph{Room Mask Condition.}
To incorporate room boundary constraints, we follow the design of DiffuScene by first resizing the binary room masks to a uniform resolution and normalizing the pixel values to form a single-channel tensor $M \in \mathbb{R}^{B \times 1 \times H \times W}$. Subsequently, a pre-trained ResNet-18 is employed as the image encoder to extract global semantic features, which are then mapped to a room embedding vector $c_{\text{room}} \in \mathbb{R}^{d_{\text{room}}}$ via a linear layer. Finally, in conjunction with the diffusion timestep $t$, $c_{\text{room}}$ is broadcast across the sequence length dimension to achieve spatially shared conditioning across the entire token sequence.

\paragraph{Text Condition.} Given a natural language description of a scene, we first employ a pre-trained BERT~\cite{devlin2019bert} model for encoding to extract a sequence of token-level semantic vectors. Subsequently, this sequence is mapped via a linear projection layer to a hidden dimension consistent with the Transformer backbone. During the denoising process, we treat the object token sequence as the Query and the encoded textual feature sequence as the Key and Value. Through the cross-attention mechanism, textual semantics are dynamically aggregated into the representation of each object token, thereby achieving the injection of textual conditions.

\subsection{Heterogeneity-aware Layout Generation}
\paragraph{Hierarchical Scene Representation.} To model indoor scenes with high density and complexity, we propose a hierarchical representation by decomposing a scene into two disjoint subsets based on semantic levels and geometric scales. 

The set of objects $\mathcal{S}^L$ consists of primary categories that form the room skeleton, with $|\mathcal{V}^L|$ classes and an upper bound of $N_L$ objects in a scene. All remaining categories are assigned to the secondary-object set $\mathcal{S}^S = \{o_i : c_i \notin \mathcal{V}^L\}$, with an upper bound of $N_S$ objects per scene. Each object $o_i$ is represented by an attribute tuple $(\mathbf{c}_i, \mathbf{t}_i, \mathbf{s}_i, \theta_i)$, where $\mathbf{c}_i \in \mathbb{R}^{C}$ is a one-hot class vector, $\mathbf{t}_i \in \mathbb{R}^{d_t}$ is the translation, $\mathbf{s}_i \in \mathbb{R}^{d_s}$ is the half-size, and the orientation angle $\theta_i$ is encoded as $(\cos\theta_i, \sin\theta_i)$. All geometric attributes are normalized to a unified reference frame. Following DiffuScene, we pad each unordered category set into fixed-slot matrices $\mathbf{X}^L \in \mathbb{R}^{N_L \times D}$ and $\mathbf{X}^S \in \mathbb{R}^{N_S \times D}$, where missing objects are filled with an empty class token. Based on this decomposition, we factorize the generation process into two heterogeneity-aware stages, SLG and CLG, enabling the two stages to collaboratively synthesize high-quality and high-density indoor scenes.

\paragraph{SLG}
In the SLG stage, the model learns the data distribution of primary categories under the joint multi-modal conditioning of the scene graph, room mask, and textual descriptions.

\paragraph{CLG}

The core design of this stage lies in implicitly learning the global dependencies between primary and secondary objects, as well as the distribution patterns of secondary objects, through the self-attention mechanism. We concatenate $\hat{\mathbf{X}}^L$ and $\mathbf{X}^S_t$ along the sequence dimension into a token sequence of length $N_L+N_S$, and apply multi-head self-attention within each Transformer block so that every secondary-object slot aggregates primary-object tokens and other secondary-object tokens under the same operator, while keeping the backbone isomorphic to SLG for reuse. $\hat{\mathbf{X}}^L$ remains a fixed anchor without diffusion noise during training and inference, and we only add noise to $\mathbf{X}^S_t$ and backpropagate gradients through the secondary-object branch, which is easier to optimize than jointly denoising all objects in a single stage.

\section{Experiments}
In this section, we conduct quantitative and qualitative evaluations of the proposed \ourmethod and prior methods on the M3dLayout dataset, aiming to verify the reliability of our approach.

\paragraph{Dataset.}
We evaluate our proposed method on M3DLayout \cite{zhangM3DLayoutMultiSourceDataset2026}, a large-scale multi-source 3D indoor layout benchmark. Specifically, we establish a set of seven types of spatial relationships to construct the scene graph representations for this dataset. The dataset integrates subsets from 3D-FRONT\cite{fu3DFRONT3DFurnished2021}, Matterport3D (MP3D)\cite{changMatterport3DLearningRGBD2017}, and the procedurally generated Inf3DLayout\cite{raistrickInfinigenIndoorsPhotorealistic2024}, covering 101 semantic categories partitioned into 64 primary object classes and 37 secondary object classes according to their functional roles in spatial structure. To improve computational efficiency in the two-stage generation pipeline and ensure physical plausibility of the synthesized scenes, only scenes satisfying $N_L < 20$ primary objects and $N_S < 100$ secondary objects are retained. After filtering, the resulting split comprises 11,508 training scenes and 1,496 validation scenes. Further implementation details can be found in the supplementary material.

\paragraph{Baselines.}
We compare our box-based layout generation approach against several state-of-the-art layout generation methods, including ATISS~\cite{paschalidouATISSAutoregressiveTransformers2021}, DiffuScene~\cite{tangDiffuSceneDenoisingDiffusion2024a}, and MiDiffusion~\cite{hu2026mixed}. To ensure a fair comparison, all baseline models are retrained and evaluated on our filtered version of the M3DLayout dataset using the same data split and BERT-based text control~\cite{devlin2019bert}. ATISS is an autoregressive layout model that generates objects and their attributes in sequence. DiffuScene is a diffusion-based layout generator that denoises layouts with a U-Net backbone. MiDiffusion is a hybrid discrete-continuous diffusion model that jointly models the noising and denoising of discrete semantic categories and continuous geometric attributes. 

\paragraph{Evaluation metrics.}
To objectively assess the plausibility of generated layouts, we follow the standard evaluation protocol widely used in 3D scene generation. For geometric evaluation, we compute the Fréchet Inception Distance (FID) and Kernel Inception Distance (KID) on 2D orthogonal projections of the predicted bounding boxes across three planes including the top-down view (XZ) and the side views (XY and YZ), thereby avoiding interference from object retrieval and surface textures in distribution-based metrics. For semantic evaluation, we use the official 3D asset library released with M3DLayout and invoke its object retrieval and rendering pipeline to convert the predicted 3D bounding boxes into top-down 2D scene images, on which we compute CLIP scores to measure consistency between the generated results and the input text. To provide a full assessment, these metrics are reported both on the full test set and across individual data sources. In the visualization of the scene after retrieval and rendering, the floor dimension is determined as the horizontal span of the rendered object. For comparisons involving multiple methods, the maximum span across all methods is adopted at the scene level to ensure consistent comparison. Further details of these metrics are provided in the supplementary material. 
\begin{table}[t]
  \caption{Quantitative comparison on the entire M3DLayout test set. KID and CLIP values are scaled by $10^{2}$ for readability. The \colorbox{colorfirst}{best} and the \colorbox{colorsecond}{second best} results are highlighted.}
  \label{tab:fid_kid_all}
  \small
  \centering
  \resizebox{\linewidth}{!}{%
  \begin{tabular}{lNNNNNNN}
    \toprule
    \textbf{Method}
      & \multicolumn{2}{c}{\textbf{XZ}}
      & \multicolumn{2}{c}{\textbf{XY}}
      & \multicolumn{2}{c}{\textbf{YZ}}
      & \textbf{CLIP} \\
    \cmidrule(lr){2-3}\cmidrule(lr){4-5}\cmidrule(lr){6-7}
      
    % \rowcolor{gray!20}
      & FID & KID & FID & KID & FID & KID & \\
    \midrule
    
    ATISS
      & 21.15 & 0.66 & 17.78 & 0.55 & 15.64 & 0.32 & 23.34 \\
    DiffuScene
      & 16.89 & \second{0.44} & \second{13.76} & \second{0.40} & 13.90 & \second{0.28} & \second{24.61} \\
    MiDiffusion
      & \second{16.71} & 0.51 & 14.90 & 0.46 & \second{13.87} & 0.30 & 24.50 \\
    \ourmethod (Ours)
      & \best{14.76} & \best{0.37} & \best{12.23} & \best{0.25} & \best{12.59} & \best{0.13} & \best{24.62} \\
    \bottomrule
  \end{tabular}
  }
  \vspace{-2mm}
\end{table}

\begin{figure*}[t] 
    \centering
    \includegraphics[width=0.9\textwidth]{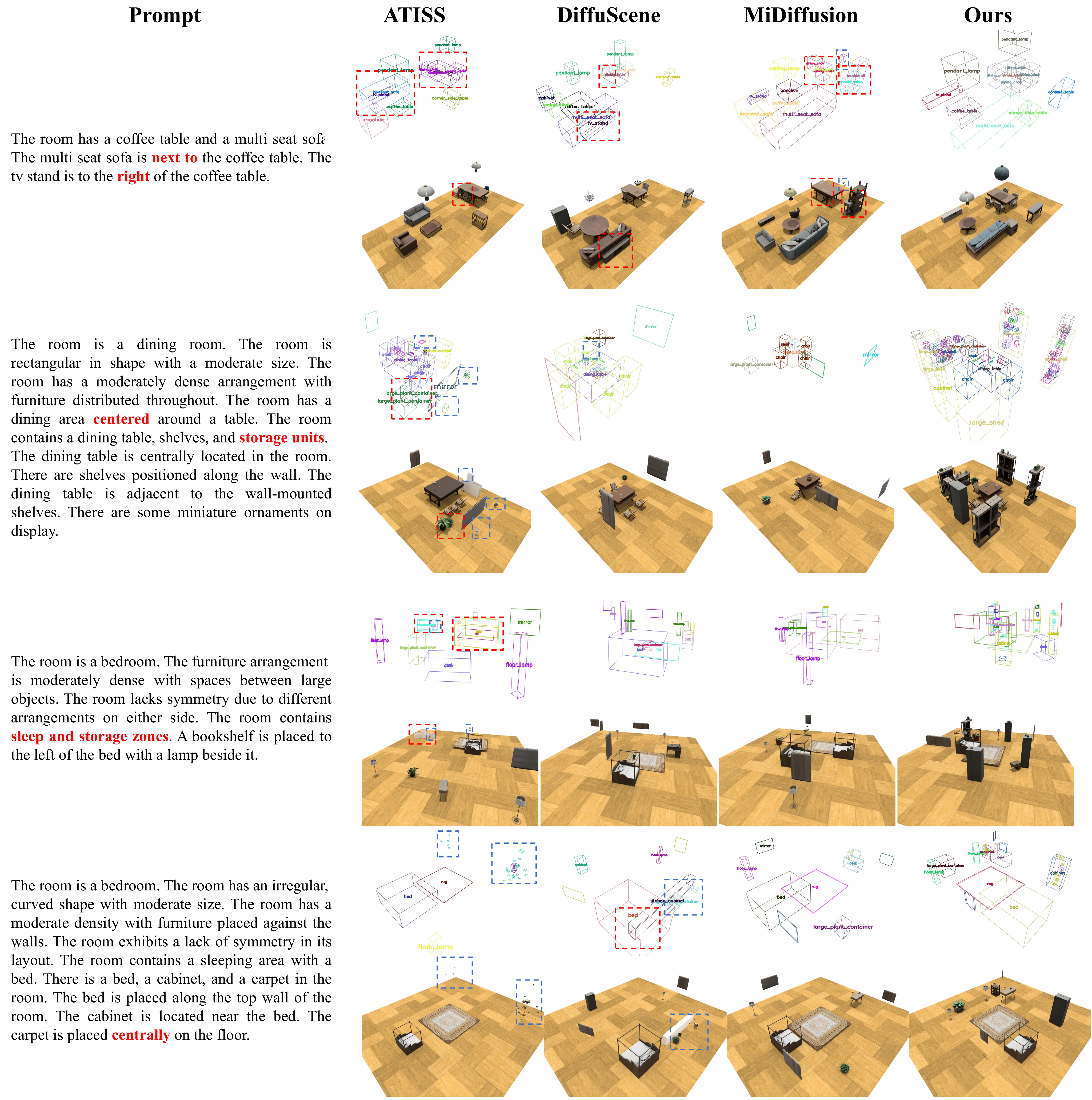}
    \caption{Comparison with SoTA scene generation models. Overlap and floating are marked with red and blue dashed boxes respectively. }
    \label{fig:compare}
    \Description{}
\end{figure*}

\begin{table*}[t]
  \caption{Quantitative results on subsets of M3DLayout test set. KID are scaled by $10^{2}$ for readability. The \colorbox{colorfirst}{best} and the \colorbox{colorsecond}{second best} results are highlighted.}
  \vspace{-0.2em}
  \label{tab:fid_kid_by_source}
  \scriptsize
  \centering
  \setlength{\tabcolsep}{1.9pt}
  \renewcommand{\arraystretch}{1.05}
  \resizebox{0.95\textwidth}{!}{%
  \begin{tabular}{lNNNNNNNNNNNNNNNNNN}
    \toprule
    \textbf{Method}
      & \multicolumn{6}{c}{\textbf{Inf3DLayout}}
      & \multicolumn{6}{c}{\textbf{MP3D}}
      & \multicolumn{6}{c}{\textbf{3DFRONT}} \\
    \cmidrule(lr){2-7}\cmidrule(lr){8-13}\cmidrule(lr){14-19}
      & \multicolumn{2}{c}{\textbf{XZ}} & \multicolumn{2}{c}{\textbf{XY}} & \multicolumn{2}{c}{\textbf{YZ}}
      & \multicolumn{2}{c}{\textbf{XZ}} & \multicolumn{2}{c}{\textbf{XY}} & \multicolumn{2}{c}{\textbf{YZ}}
      & \multicolumn{2}{c}{\textbf{XZ}} & \multicolumn{2}{c}{\textbf{XY}} & \multicolumn{2}{c}{\textbf{YZ}} \\
    \cmidrule(lr){2-3}\cmidrule(lr){4-5}\cmidrule(lr){6-7}
    \cmidrule(lr){8-9}\cmidrule(lr){10-11}\cmidrule(lr){12-13}
    \cmidrule(lr){14-15}\cmidrule(lr){16-17}\cmidrule(lr){18-19}
      & FID & KID & FID & KID & FID & KID
      & FID & KID & FID & KID & FID & KID
      & FID & KID & FID & KID & FID & KID \\
    \midrule
    ATISS
      & 38.08 & 1.64 & 30.65 & 1.13 & 29.01 & 0.76
      & 72.84 & 0.86 & 69.38 & 1.42 & 63.18 & 1.06
      & 20.71 & 0.34 & \second{18.44} & \second{0.57} & \second{15.48} & \second{0.17} \\
    DiffuScene
      & 30.34 & 1.05 & 21.69 & \best{0.39} & \best{22.92} & \best{0.28}
      & \best{63.94} & \best{0.40} & 55.83 & \best{0.56} & 60.97 & \second{0.86}
      & \second{20.06} & \second{0.28} & 19.89 & 0.62 & 18.91 & 0.41 \\
    MiDiffusion
      & \second{27.07} & \second{0.89} & \second{21.47} & \second{0.42} & 24.12 & 0.38
      & \second{67.41} & \second{0.72} & 67.30 & 1.22 & \second{60.13} & 1.15
      & 24.17 & 0.49 & 30.11 & 1.45 & 27.28 & 0.95 \\
    \ourmethod (Ours)
      & \best{23.73} & \best{0.59} & \best{21.25} & \second{0.45} & \second{23.09} & \second{0.30}
      & 79.10 & 2.28 & \best{53.65} & \second{0.62} & \best{53.98} & \best{0.31}
      & \best{16.72} & \best{0.18} & \best{14.64} & \best{0.12} & \best{14.92} & \best{0.11} \\
    \bottomrule
  \end{tabular}%
  }
  \vspace{-1mm}
  
\end{table*}
\paragraph{Implementation details.}
The initial learning rate is set to $10^{-4}$ and halved every 10{,}000 training epochs. The model is trained for 30{,}000 epochs with batch size 64. Following DiffuScene, we adopt a linear noise schedule ranging from 0.0001 to 0.02 over 1{,}000 diffusion steps. Gradient clipping is applied with a threshold of 10. All experiments are conducted with two servers equipped with 2 NVIDIA GeForce RTX 4090 and 4 A800 GPUs, respectively.

\subsection{Comparison with other methods}
We compare \ourmethod both qualitatively and quantitatively with current mainstream scene generation models, including DiffuScene, MiDiffusion, and ATISS.

Table \ref{tab:fid_kid_all} presents the evaluation results on the full test set. It can be observed that \ourmethod consistently outperforms all baseline models across all three orthogonal projection planes. Specifically, our model achieves the lowest FID and KID scores, with a significant improvement in FID on the XZ plane—representing the floor plan layout—where it reduces the score by approximately 11.7\% compared to the second-best model. This demonstrates that \ourmethod more effectively learns the joint distribution aligned with real-world scenes, thereby maintaining a robust advantage in both top-down and dual side-view evaluations. Regarding text-semantic consistency, the CLIP scores indicate that while significantly enhancing structural layout quality, \ourmethod fully preserves the capacity for text-to-scene alignment.

To further verify the robustness of our model, we report the performance breakdown by data source in Table \ref{tab:fid_kid_by_source}. In the 3DFRONT dataset, characterized by strong layout logic, and the Inf3DLayout dataset, featuring dense distributions of secondary objects, \ourmethod achieves the best overall performance across all three viewing angles. Although the lead in the XZ view is narrowed compared to the side views on the MP3D dataset, \ourmethod maintains a consistent advantage in side-view metrics. This discrepancy primarily stems from the inherent noise in the real-world scanned data of MP3D, as well as the domain shift resulting from the relatively limited sample scale within the dataset.

The qualitative comparison in Fig.~\ref{fig:compare} further substantiates the superiority of \ourmethod in spatial reasoning and structural integrity. While baseline models frequently suffer from issues such as floating objects or structural overlaps, \ourmethod produces physically plausible layouts with increased complexity. These visual results align with improvements in metrics, demonstrating that our approach remains superior in learning the data distributions of both primary objects and secondary objects.

\begin{figure}[t]
  \centering
  \includegraphics[width=0.85\linewidth]{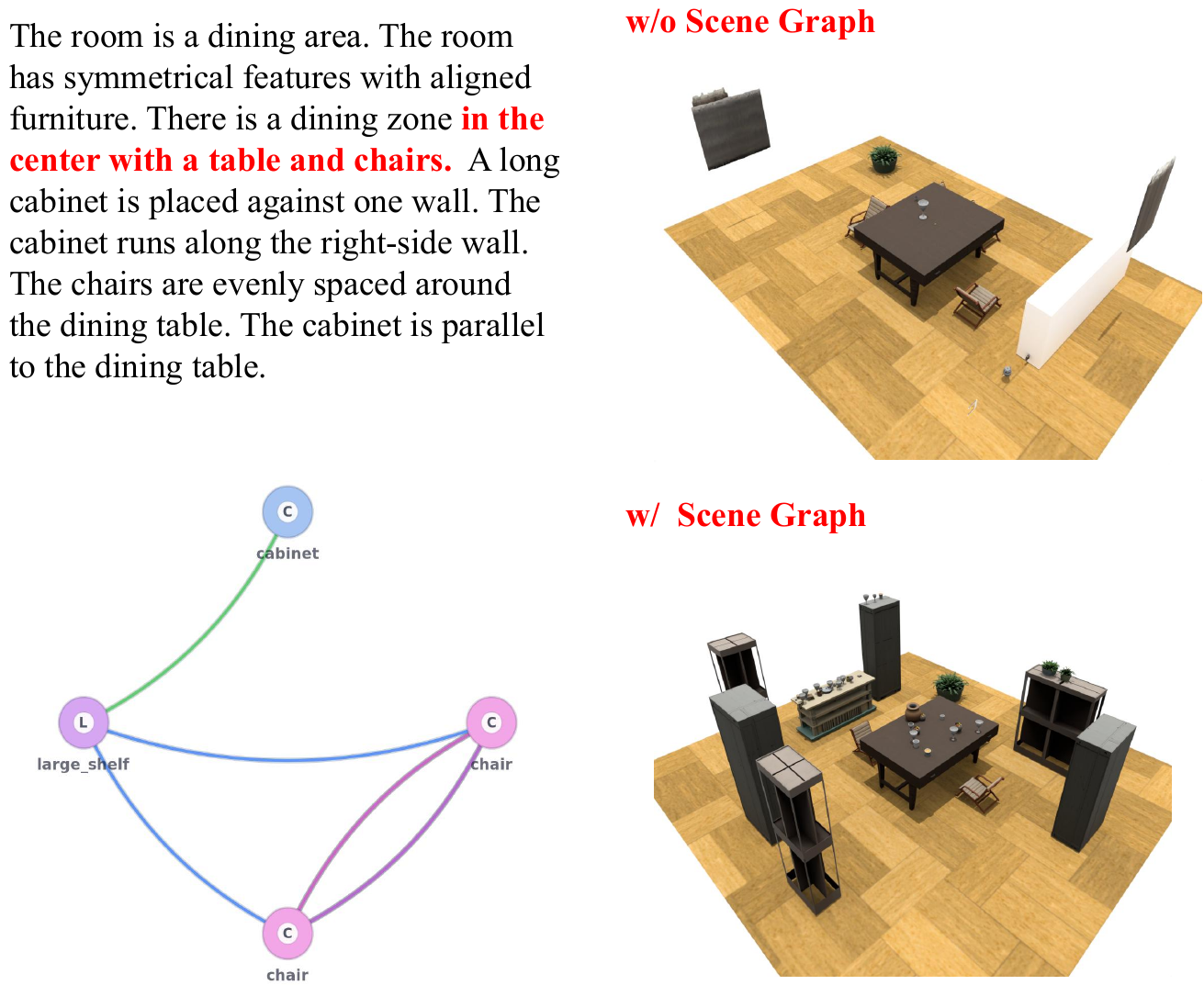}
  \caption{Qualitative ablation of LSG condition.}
  \Description{Top: model without LSG condition. Bottom: model with LSG condition.}
  \label{fig:ablation}
  \vspace{-2mm}
  
\end{figure}

\begin{table}[t]
  \caption{Ablation on the entire M3DLayout test set. KID and CLIP values are scaled by $10^{2}$ for readability. The \colorbox{colorfirst}{best} and the \colorbox{colorsecond}{second best} results are highlighted. BS: the baseline 1D-UNet denoising network; TB: the Transformer-based denoising backbone with learnable spatial-semantic modulation; HP: heterogeneity-aware layout generation pipeline; LC: the LSG condition.}
  \label{tab:ablation_fid_kid_all}
  \small
  \centering
  \resizebox{\linewidth}{!}{%
  \begin{tabular}{lNNNNNNN}
    \toprule
    \textbf{Method}
      & \multicolumn{2}{c}{\textbf{XZ}}
      & \multicolumn{2}{c}{\textbf{XY}}
      & \multicolumn{2}{c}{\textbf{YZ}}
      & \textbf{CLIP} \\
    \cmidrule(lr){2-3}\cmidrule(lr){4-5}\cmidrule(lr){6-7}

      & FID & KID & FID & KID & FID & KID & \\
    \midrule
    BS
      & 16.89 & 0.44 & \second{13.76} & 0.40 & \second{13.90} & 0.28 & \second{24.61} \\
    BS + TB
      & 16.20 & \second{0.40} & 15.21 & 0.64 & 14.65 & 0.37 & 24.57 \\
    BS + TB + HP
      & \second{15.59} & 0.42 & 13.86 & \second{0.40} & 13.93 & \second{0.26} & 24.48 \\
    BS + TB + HP + LC (Ours)
      & \best{14.76} & \best{0.37} & \best{12.23} & \best{0.25} & \best{12.59} & \best{0.13} & \best{24.62} \\
    \bottomrule
  \end{tabular}
  }
  \vspace{-2mm}
\end{table}

\subsection{Ablation study}

As summarized in Table $\ref{tab:ablation_fid_kid_all}$, the progressive ablation results demonstrate the efficacy of each proposed component in enhancing layout quality. The transition from the convolutional baseline to the Transformer-based architecture yields a marked improvement in metrics across the XZ perspective, which confirms the superiority of the global self-attention mechanism in modeling long-range dependencies among objects. However, the alignment in side views remains less sensitive due to the scale shift between primary and secondary objects. By introducing the heterogeneity-aware layout generation pipeline (HP) to decouple structural layout optimization from fine-grained entity synthesis, the optimization competition induced by scale variances is effectively mitigated. Building upon this, our full model further integrates the Learnable Scene Graph (LSG) condition to encode explicit topological constraints as cross-attention context, with visual comparisons shown in Fig. $\ref{fig:ablation}$. It is observed that by explicitly introducing geometric prior conditions, the spatial logic and complexity of the generation results are enhanced. Meanwhile, the CLIP scores remain stable across all configurations, indicating that this work enhances the physical plausibility and complexity of the layouts while effectively maintaining the semantic consistency of the scenes.

% \subsection{More result.}

% We provide additional qualitative results generated by HetScene in Fig.~\ref{fig:res_more}. These samples further validate the efficacy of our model in synthesizing complex indoor environments with high structural fidelity and physical plausibility.

\section{Conclusion}
In this work, we propose \ourmethod, a two-stage hierarchical diffusion framework for generating high-quality simulation-ready scenes for embodied intelligence. The scene synthesis task is decomposed into \emph{Structural Layout Generation}~(SLG) and \emph{Contextual Layout Generation}~(CLG). In SLG, we introduce a \emph{Learned Scene Graph} condition that encodes pairwise spatial relations into compact conditioning signals to improve global structural and relational consistency. In CLG, the model anchors on the primary objects generated in the first stage, models dependencies between primary objects and secondary objects via concatenated-sequence self-attention, thereby maintaining global layout plausibility while enriching fine-grained details locally. Experiments on the M3DLayout benchmark show consistent improvements in layout quality, object diversity, and scene density. We believe that the principle of semantic-level decomposition adopted in this work is broadly applicable and can be extended to other layout and structured generation tasks with hierarchical organization.
% \bigskip

% \noindent \lipsum[1] \cite{1}

$\,$

$\,$

% \begin{thebibliography}{99}

% \bibitem{1} Spiegel, M. R. (1981). Theory and problems of Advanced Calculus: Si (metric) edition. McGraw-Hill. 

% \end{thebibliography}
% \bibliographystyle{unsrt} 
% \bibliography{reference}
% \bibliographystyle{ACM-Reference-Format}
% \bibliography{sample-base}

\printbibliography
\end{document}